\documentclass[10pt,pra,aps,twocolumn,amsmath,amssymb,superscriptaddress]{revtex4-1}

\usepackage{times}
\usepackage{textcomp}
\usepackage{graphicx}
\usepackage[caption=false]{subfig}
\usepackage{bm}
\allowdisplaybreaks 
\usepackage{braket}

\makeatletter

\newcommand{\Rmnum}[1]{\expandafter\@slowromancap\romannumeral #1@}
\makeatother

\usepackage[breaklinks]{hyperref}
\hypersetup{colorlinks=true, linkcolor=blue, citecolor=blue, filecolor=blue, urlcolor=blue}

\begin{document}

\title{A Multi-Scale Tensor Network Architecture for Classification and Regression}

\author{Justin Reyes}
\affiliation{Department of Physics, University of Central Florida, 4000 Central Florida Blvd, Orlando, FL 32816, USA}

\author{E. Miles Stoudenmire}
\affiliation{Center for Computational Quantum Physics, Flatiron Institute, 162 5th Avenue, New York, NY 10010, USA}

\date{\rm\today}

\begin{abstract}
    We present an algorithm for supervised learning using tensor networks,
    employing a step of preprocessing the data by coarse-graining through
    a sequence of wavelet transformations. We represent these transformations
    as a set of tensor network layers identical to those in a
    multi-scale entanglement renormalization ansatz (MERA) tensor network, 
    and perform supervised learning and regression tasks through a model 
    based on a matrix product state (MPS) tensor network acting on the 
    coarse-grained data. Because the entire model consists of tensor contractions
    (apart from the initial non-linear feature map), we can adaptively fine-grain the
    optimized MPS model backwards through the layers with essentially no loss in performance. 
    The MPS itself is trained using an adaptive algorithm based 
    on the density matrix renormalization group (DMRG) algorithm. 
    We test our methods by performing a classification task on audio data and a regression 
    task on temperature time-series data, studying the dependence of training accuracy on 
    the number of coarse-graining layers and showing how fine-graining through the network
    may be used to initialize models with access to finer-scale features.
\end{abstract}

\maketitle

\section{Introduction}
\label{sec:intro}

Computational techniques developed across the machine learning and physics fields
have consistently generated promising methods and applications in both areas of study. 
The application of well established machine learning architectures and optimization techniques has enriched the physics community with advances such as modeling and recognizing topological 
quantum states~\cite{Marcello2019,Chinni2019,Rodriguez:2019}, 
optimizing quantum error correction codes~\cite{Nautrup2018}, or
classifying quantum walks~\cite{Melkinov2019}.

Conversely, techniques known as tensor networks which model high-dimensional functions 
and are closely connected to physical principles have begun to be explored more in 
applied mathematics and machine learning ~\cite{Novikov:2015aa,Novikov:2016,StoudenmireJan2017,StoudenmireDec2017,
Glasser2018,Yu2018,Han:2018,Cheng2019,Levine2017,Glasser:2019,Batselier:2019}.
Benefits of the tensor network approach include compression of model parameters~\cite{Novikov:2015aa}, adaptive training algorithms \cite{StoudenmireJan2017}, and the possibility
to gain theoretical insights \cite{Glasser:2019,Bradley:2019}.
In this paper, we focus on the application of tensor networks the machine learning
setting, seeking to explore what are some of the interesting capabilities tensor network techniques could provide.

A tensor network is a factorization of a very high-order tensor---a tensor
with a large number of indices---into a set of low-order tensors whose indices
are summed to form a network defined by a certain pattern of contractions. 
Such a network can be viewed as a generalization of low-rank matrix
decompositions familiar in applied mathematics. 
Through a tensor network decomposition, the number of parameters needed to model the
high-order tensor can be exponentially reduced while still approximating the high-order 
tensor accurately within many applications of interest \cite{Orus2014}.

One setting where tensor networks serve as a foundational tool is in the simulation of 
quantum many-body systems, where the system state vector---which encodes
the joint probability of many variables---cannot be explicitly 
stored or manipulated because of its exponentially growing dimension 
with the number of variables. 
Certain machine learning approaches, such as kernel learning, 
 encounter this same problem when a model involves 
a large number of data features~\cite{Cichocki2014}. 
The similarity in the growth of the dimensionality of vector spaces in the 
two fields of study motivates the use of tensor network algorithms for enhancing the performance of machine learning tasks. 

\begin{figure}[th]
\centering
\includegraphics[width=0.6\columnwidth]{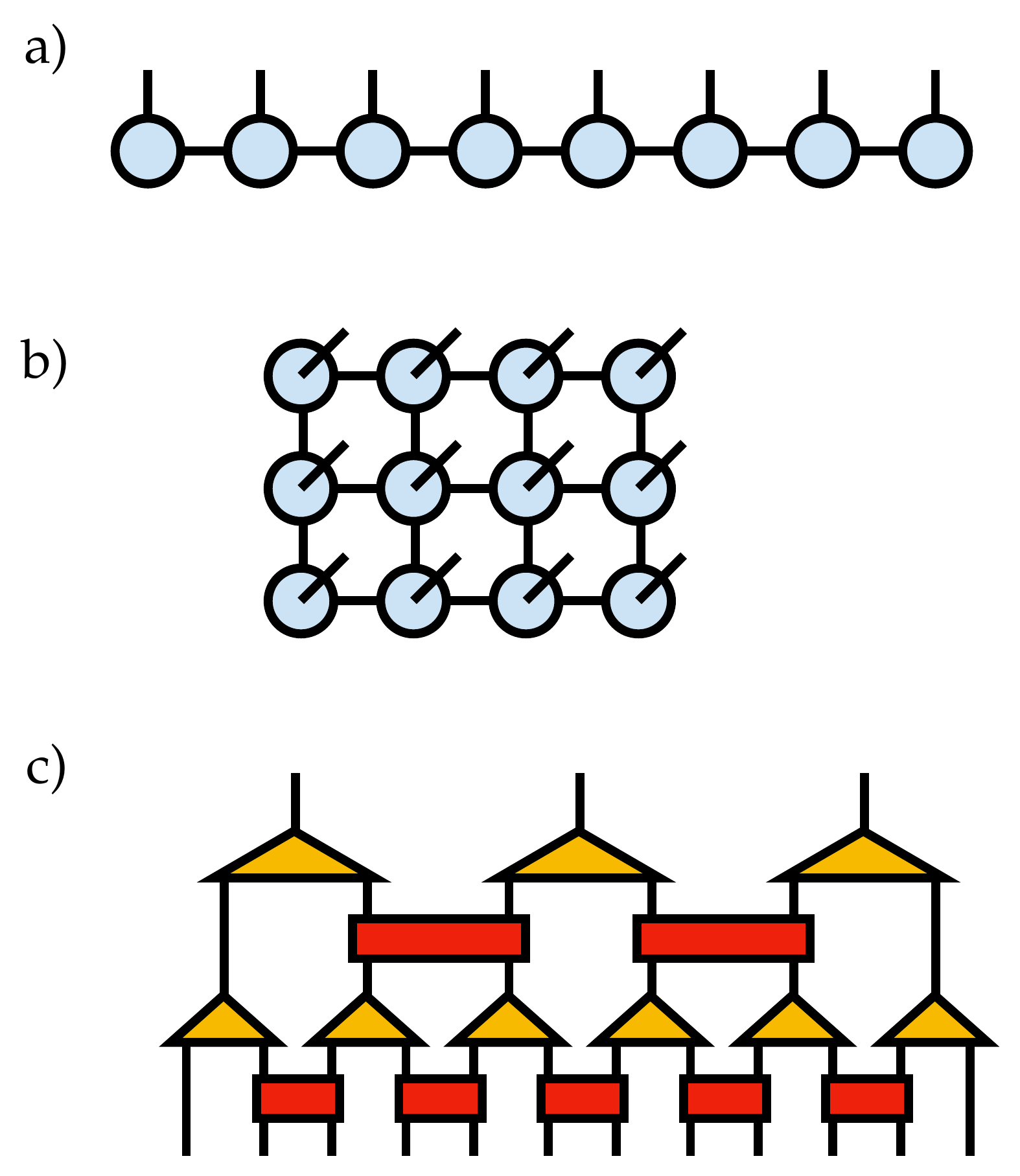}
\caption{Tensor diagrams for the a) matrix product state (MPS) 
tensor network, b) PEPS tensor network, and c) two-layer MERA tensor network.} 
\label{fig:tensor_diagrams}
\end{figure}

Matrix product states (MPS)
or tensor trains~\cite{Vidal:2003,Perez-Garcia:2007,Schollwock2011,Oseledets:2011}, 
projected entangled pair states (PEPS)~\cite{Orus2014}, and the 
multi-scale renormalization ansatz (MERA)~\cite{Vidal2008} are all common tensor 
network families with well developed optimization algorithms \cite{Evenbly2009}. 
Example tensor diagrams defining each of the above mentioned architectures are 
shown in Fig.~\ref{fig:tensor_diagrams}.
Because tensors are multi-linear, controlled transformations of one tensor network
into another are possible through techniques such as matrix factorization and
tensor contraction \cite{Vidal:2007algv2,Dolfi:2012,Batselier:2019}. 
We will take advantage of the capability of one tensor network to transform into another
network to initialize our model, by initially training a model defined through a
larger number of coarse-graining steps, then fine-graining the top tensor 
containing the adjustable model weights to reach the desired architecture.

The algorithm we present for training a model for supervised learning and regression
applications is built around three main steps. The first step involves coarse graining 
to reduce the feature space representing the data, and consequently the number of parameters of the model. This is done through a series of wavelet transformations approximated by 
the layers of a MERA tensor network. The second step is the training of a weight tensor represented as an MPS and optimized by an alternating least squares algorithm 
analogous to the density matrix renormalization group (DMRG) algorithm used in physics. 
In the third step, we show how the weight tensor
can be controllably fine-grained into a model over a larger set of features, which initially has the same performance as the coarse-grained model but can then be further optimized.

This paper is organized as follows: In Section \ref{sec:background}, we provide 
a high-level overview of wavelet transformations, 
the DMRG algorithm, and the MERA tensor network. In Section \ref{sec:algorithm}, 
we detail the methods involved in encoding the MERA with wavelet layers, and describe our algorithm for training the weights of the classifier. In Section \ref{sec:results}, we apply our method to the classification of DCASE audio file data sets and the linear regression of mean temperature data sets. We conclude in Section \ref{sec:discussion} with a discussion on the applicability and outlook of our algorithm.

\section{Theoretical Background}
\label{sec:background}
The steps of the algorithm we will explore combine a number of techniques from 
different fields of study. In this section we present a brief review 
of these techniques to aid readers not familiar with some of them and to show
why we are motivated to combine them together. Previous works in 
the physics literature have been similarly motivated to adapt multi-scale or 
multi-grid modeling ideas into the tensor network setting \cite{Vidal:2007algv2,Dolfi:2012}.

In the following section, we present background for the machine learning tasks
we will discuss; review the theory of wavelet transformations; 
and review the MPS tensor network, DMRG algorithm; 
and the MERA tensor network.
 
\subsection{Learning Task and Model Functions}

We will be interested in the tasks of regression and   supervised learning. 
Given a training data set of observed input and output pairs 
$\{\mathbf{x}_j, \mathbf{y}_j\}_{j=1}^{n}$, both of these
tasks can be formulated as finding a function $f(\mathbf{x})$ satisfying
$f(\mathbf{x}_j) \approx \mathbf{y}_j$ to a good approximation. 
Crucially, out of all functions which fit the input data, $f(\mathbf{x})$ should be 
selected so as to generalize to unobserved data.
Outputs $\mathbf{y}_j$ can be vector-valued, but in the case of 
regression are more commonly just scalar-valued.

The class of model functions we will optimize have the general form
\begin{align}
\label{eq:decision_f}
    f_{W}(\mathbf{x}) = W \cdot \Phi(\mathbf{x})
\end{align}
where $W$ is the weight vector which enters linearly. The function $\Phi$ is 
the feature map, which is generally non-linear and maps the input $\mathbf{x}$ to a feature vector $\Phi(\mathbf{x})$.
This class of model functions is formally the same starting point 
as the kernel learning and support vector machine approaches to machine learning.
Unlike these approaches, we will optimize over the weights $W$ without introducing any dual parameters as in the kernel trick. To make such an optimization feasible, 
we represent the weights in a compressed form as a tensor network.

The feature maps we consider take the form
\begin{align}
    \Phi^{s_1 s_2 \cdots s_N}(\mathbf{x}) = \phi^{s_1}(x_1) \phi^{s_2}(x_2) \cdots \phi^{s_N}(x_N)
\end{align}
where $N$ is the dimension of the input vectors $\mathbf{x}$,
and each index $s_i$ runs over $d$ values. The \emph{local feature maps} 
$\phi$ thus map each input component $x_i$ to a $d$-dimensional vector.
The resulting feature map $\Phi$ is a map from $\mathbb{R}^N$ to a rank-1 tensor
of order $N$, whose indices are all of dimension $d$. Informally, $\Phi$
is a product of vector-valued functions of each of the input components.

Our goal is to find a suitable weight vector $W^*$ 
that minimizes some appropriate cost function between the 
model output $f_{W^*}(\mathbf{x}_j)$ and the expected output $\mathbf{y}_j$ over the observed data. 
The cost function we will use is the quadratic cost
\begin{equation}\label{eq:cost}
    C(W) = \frac{1}{2 n}\sum_{j=1}^n{||f_{W}(\mathbf{x}_j) - \mathbf{y}_j||^2} + \lambda ||W||^2
\end{equation} 
where the summation is over $n$ training examples and $\lambda$ is an empirical
regularization parameter.
Because of the quadratic form of this function, optimization can be conveniently 
carried out by efficient methods such as the conjugate gradient algorithm.   

\subsection{Wavelet Transformations}
In signal processing, the Fourier transform is a ubiquitous transformation which 
can be used to take an input signal from the time domain to the frequency domain. 
However, there are many instances where a signal is not stationary, requiring 
analysis to be done in both the time and the frequency domain. Wavelet 
transformations facilitate this type of analysis by transforming a signal into the 
time-frequency domain \cite{Walker1999}. 

The continuous wavelet transformation of an input signal $x(t)$ is given by
\begin{equation}\label{eq:cwt}
X(\tau,s) = \frac{1}{\sqrt{|s|}}\int x(t) \bar{\psi}\left(\frac{t-\tau}{s}\right) dt
\end{equation}   
where $\psi$ is the \textit{mother wavelet} 
which characterizes the transformation, $\tau$ is the translation parameter, 
and $s$ is the scale parameter.

For discrete signals taken over constant time intervals, the mother wavelet becomes
 a vector whose inner product is taken with a subset of the signal $x$. The 
integral becomes a summation over the elements within that subset; neighboring
subsets can generally overlap.
 
The two types of discrete wavelet transformations we will use below are 
Haar transformation
\begin{equation}\label{eq:h2}
h^{(2)}_i = \frac{x_{2i} + x_{2i+1}}{\sqrt{2}} 
\end{equation} 
for $i=0$ to $i=N/2-1$ 
and the Daubechies-4 (Daub4) transformation 
\begin{equation}\label{eq:d4}
d^{(4)}_i = \sum_{j=0}^3{x_{2i+j}D_j}
\end{equation}
where the coefficients $D_j$ are the elements of the vector
\begin{equation}\label{eq:d4D}
D = \left( \frac{1+\sqrt{3}}{4\sqrt{2}},\frac{3+\sqrt{3}}{4\sqrt{2}},\frac{3-\sqrt{3}}{4\sqrt{2}},\frac{1-\sqrt{3}}{4\sqrt{2}} \right) \ .
\end{equation}

For both of these discrete transformations the translation parameter is $\tau = 2$.
The summation is over two elements for the Haar transformation and four elements 
for the Daub4 transformation.  
These discrete wavelet transformations of an input signal $x$ can be used to average and rescale the initial signal from size $N$ 
to size $N/2$, while preserving local information in both the time and frequency domains. The information that is preserved is that associated with the part of the signal which varies more smoothly with the time index $i$.

\subsection{MPS Tensor Network and Optimization}

A matrix product state (MPS) or tensor train is a decomposition of a high order tensor into a 
contracted network order-3 tensors
with a one-dimensional, chain-like structure as in Fig.~\ref{fig:tensor_diagrams}(a). 
In traditional tensor notation an MPS decomposition of a tensor $W^{s_1 s_2 \ldots s_N}$
can be written as
\begin{equation}
W^{s_1 s_2 s_3 \ldots s_N} 
= \sum_{\{a\}} A^{s_1}_{a_1} A^{s_2}_{a_1 a_2} A^{s_3}_{a_2 a_3} \cdots A^{s_N}_{a_{N-1}}
\end{equation}
where the factor tensors $A^{s_j}_{a_{j-1} a_j}$ can in general be different from each other.

The key parameter controlling the expressivity of an MPS is the \emph{bond dimension},
also known as tensor-train rank. This is the dimension of the internal, contracted 
indices $\{a_j\}$ connecting neighboring factor tensors of the MPS. 
By taking the bond dimension large enough, an MPS can represent any tensor \cite{Vidal:2003,Perez-Garcia:2007}. In general, the dimensions of the bond indices vary, in which
case the bond dimension of the MPS as a whole means the maximum over the 
bond dimensions.

There are many important theoretical results about MPS, such as finding minimal sets
of conditions sufficient to make the parameters of the factor tensors uniquely 
specified \cite{Perez-Garcia:2007,Schollwock2011,Oseledets:2011}.
For our purposes, the most important fact about MPS will be that they
are well-suited for modeling functions or processes with strongly one-dimensional 
correlations, such as samples of a time-dependent random variable with decaying correlations.
MPS can exactly reproduce long-time correlations which decay exponentially, and
can also approximate power-law correlations to high accuracy \cite{Evenbly2011}.

A straightforward but powerful way to optimize a MPS for a given objective
is to optimize each of the factor tensors $A^{s_j}_{a_{j-1} a_j}$ one-by-one, keeping
the others fixed. In physics, the adaptive version of this approach is generally 
referred to as the DMRG algorithm \cite{White1992}, while in mathematics it is 
known as alternating least squares (ALS). 
By sweeping from the first tensor to the last and then back when optimizing, 
computations involving the other factor tensors can be reused, making the approach
highly efficient. Another advantage of alternating optimization is that it can be 
made adaptive by temporarily contracting over the bond index shared between two factor 
tensors. 
After optimizing the resulting tensor, the MPS form can be 
restored by an singular value factorization, introducing a new bond index that can
be selected larger or smaller depending on the desired tradeoff between the quality of the 
results and the computational cost \cite{StoudenmireJan2017}.

\subsection{MERA}

\begin{figure}[b]
\centering
\includegraphics[width=\columnwidth]{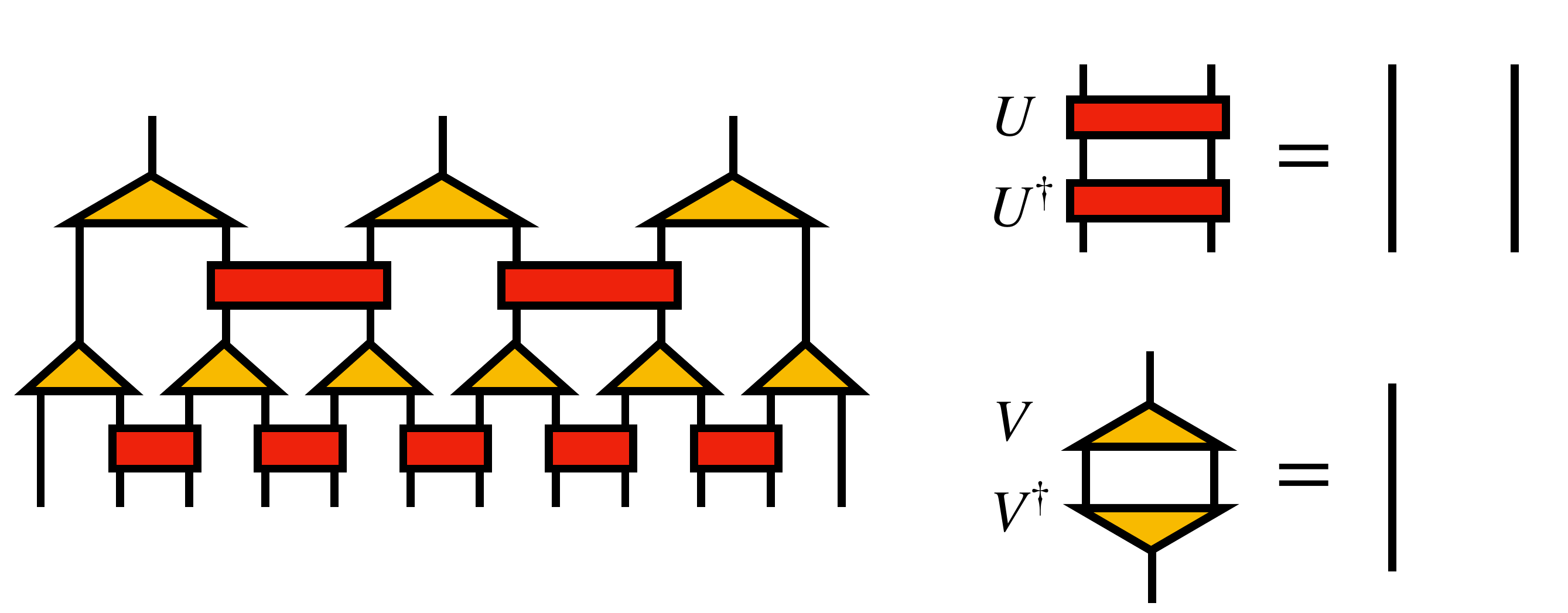}
\caption{Two layers of a MERA tensor network, showing the unitary condition
obeyed by the disentanglers $U$ and isometric condition obeyed by the isometries $V$.} 
\label{fig:MERA}
\end{figure}

The MERA is a tensor network whose geometry and structure implements multiple 
coarse grainings of input variables. In physics terminology, such a coarse-graining 
process is known as renormalization~\cite{Wilson:1979}. While a coarse-graining
process could be implemented by a tree tensor network, a tree has the drawback that
information or features processed by different subtrees are not merged until the
subtrees meet, which can happen at an arbitrarily high scale. The MERA architecture
fixes this problem by including extra ``disentangler'' tensors which span across
subtrees \cite{Vidal2007,Evenbly2009}---these are the four-index tensors $U$ shown in Fig.~\ref{fig:MERA}.

Introducing tensors which connect subtrees
could make computations with the resulting network prohibitively expensive. 
However, in a MERA the disentangler tensors $U$ are constrained to always be unitary with 
$U^\dagger U = U U^\dagger = 1$.
Likewise, the tree tensors, or isometry tensors $V$ are constrained to obey an
isometric condition $V^\dagger V = 1$ (yet $V V^\dagger \neq 1$). These conditions
are depicted in diagrammatic form in Fig.~\ref{fig:MERA}. Because of these constraints,
computations of the norm of a tensor represented by a MERA, or of marginals and
correlation functions when the MERA represents a distribution, can be carried out
with a cost scaling polynomially in the dimensions of the internal indices of the network.

In this work, we will specifically consider disentanglers and isometries 
parameterized as:
\[
U = 
\begin{pmatrix}
    1 & 0 & 0 & 0\\
    0 & \cos \theta_U & \sin \theta_U & 0\\
    0 & -\sin \theta_U & \cos \theta_U & 0\\ 
    0 & 0 & 0 & 1
\end{pmatrix}
\]  
\[
V = 
\begin{pmatrix}
    1 & 0 & 0 & 0 \\
    0 & \sin \theta_V & \cos \theta_V & 0 
\end{pmatrix}
\]
which will be sufficient to parameterize MERA layers which approximately compute wavelet coarse
graining transformations of input data. 
We discuss how to choose the angles $\theta_U$ and $\theta_V$
to approximate Haar and Daubechies wavelets in the next section.
The existence of a correspondence between the MERA tensor network and discrete
wavelet transformations was first described in Refs.~\onlinecite{Evenbly:2016,Evenbly:2018w}.

\section{Model and Training Algorithm}

The model function we will now discuss for regression and supervised learning first coarse grains input data through some number of discrete wavelet transformations, implemented as MERA tensor network layers. Then an MPS tensor network is used to represent the top layer of trainable weights. 

After discussing how to train a model of this type, we highlight one of its key advantages: the amount of coarse graining can be adjusted \emph{during training} to adaptively discover the number of coarse graining steps needed to obtain satisfactory results.

\label{sec:algorithm}
\begin{figure}[th]
\centering
\includegraphics[width=\columnwidth]{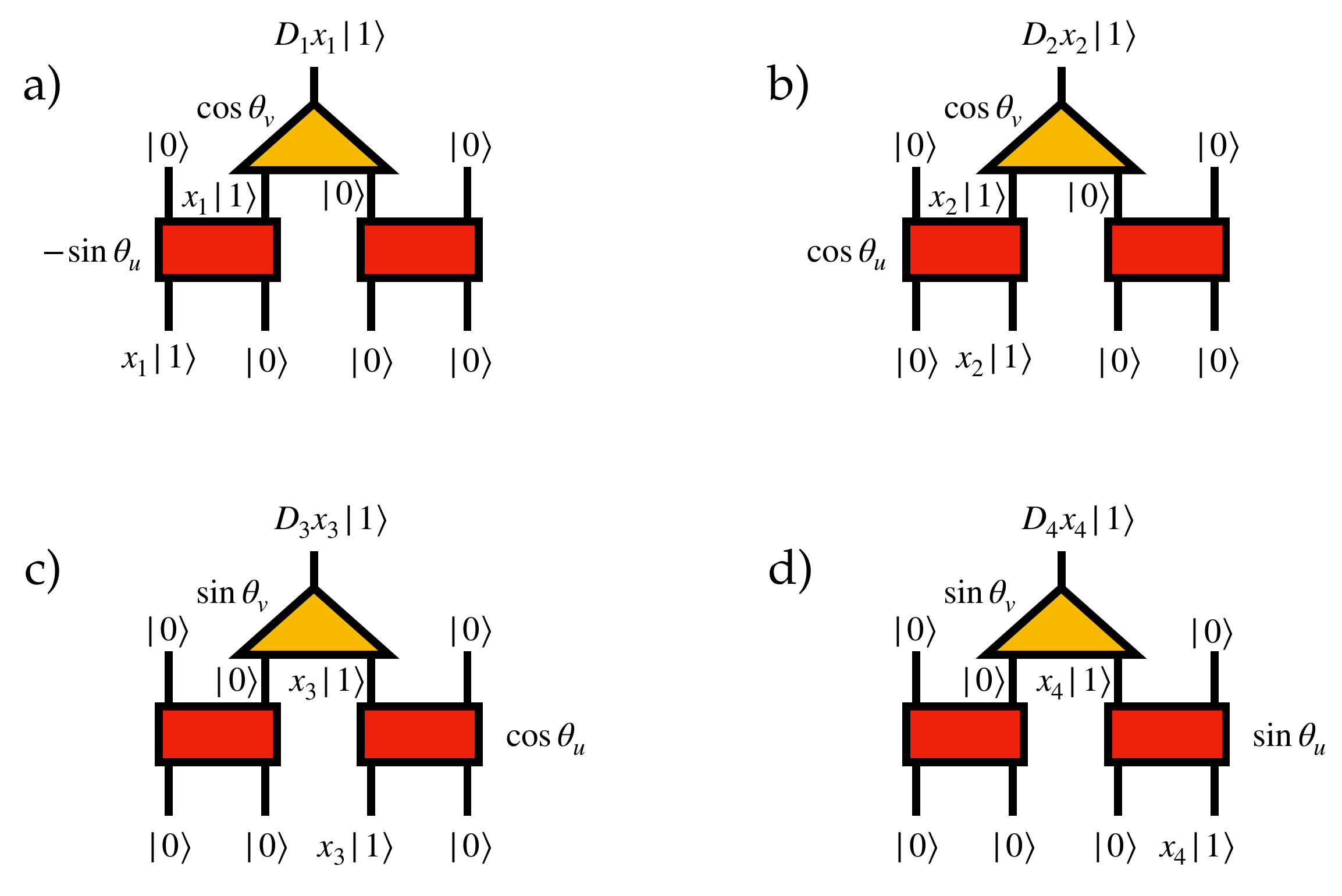}
\caption{Decomposition of each transformation element $D_{1-4}$ into the unitaries and isometries of the wavelet MERA. The complete Daub4 wavelet transformation over data elements $x_{1-4}$ is given by the contraction over connected indices. Each transformation is parameterized by $\theta_U$ and $\theta_V$. } 
\label{fig:wav_mera}
\end{figure}
\begin{figure}[th]
\centering
\includegraphics[width=\columnwidth]{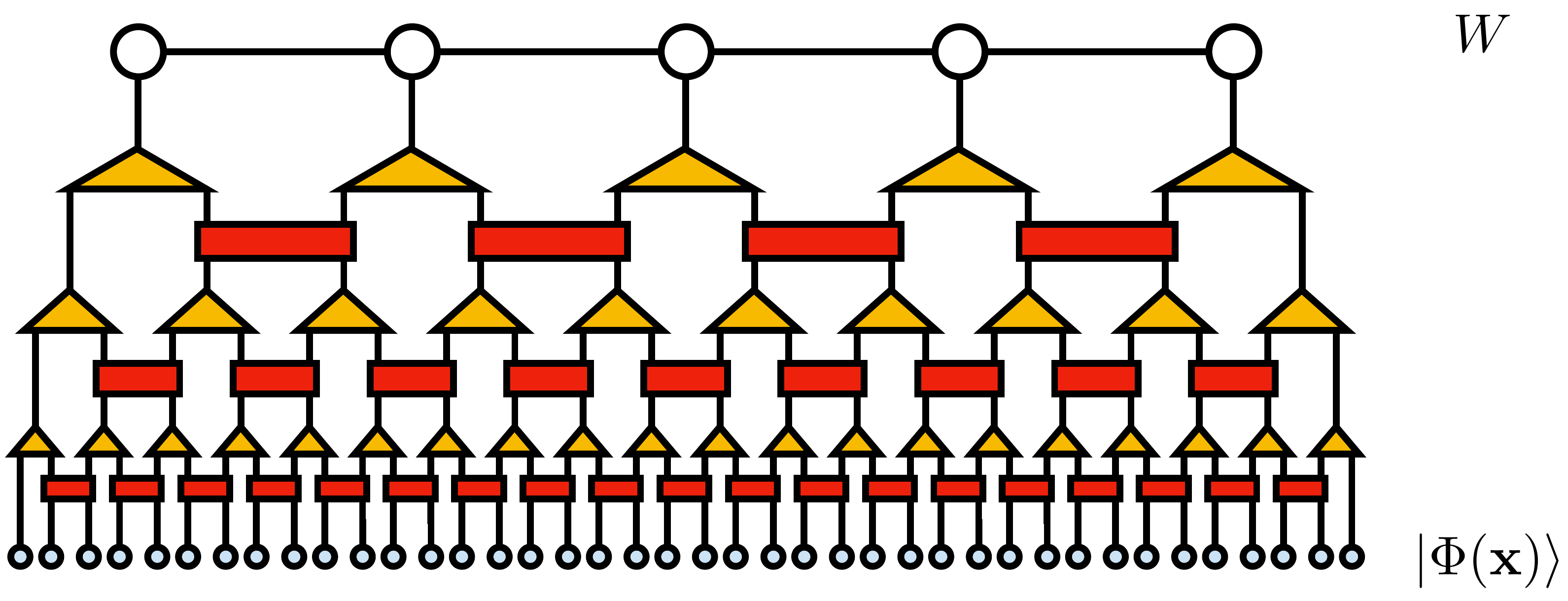}
\caption{The tensor diagram for the model we use for classification and regression, showing
the case of three MERA layers. 
Each input data $\mathbf{x}$ is first mapped into a (rank-1) MPS $\ket{\Phi(\mathbf{x})}$, 
then acted on by MERA layers approximating Daub4 wavelet transformations. At the top layer, the
trainable parameters of the model $W$ are decomposed as an MPS. Because all
tensor indices are contracted, the output of the model is a scalar.} 
\label{fig:wav_mera_diagram}
\end{figure}

\subsection{Coarse Graining}

To reduce the parameter space necessary for training the weights our classifier, we coarse grain the input data through a series of wavelet transformations, effectively reducing the size of the data by a factor of two after each transformation. This is done by first mapping 
each input data element $x_i$ to the vector $\ket{\phi(x_i)} = \ket{0} + x_i\ket{1}$, 
where in this section we use the physics notation that $\ket{v}$ is a vector labeled $v$. 
We have also defined
\begin{align}
\ket{0} = (1\ 0)^{T} \\
\ket{1} = (0\ 1)^{T} 
\end{align}
The feature map applied to each data sample is taken to be 
the tensor product  
\begin{equation}\label{psi_x}
\ket{\Phi(\mathbf{x})} = \ket{\phi(x_1)}\otimes\ket{\phi(x_2)}\otimes ...\otimes\ket{\phi(x_N)} \ .
\end{equation}
where the $\otimes$ symbol is often omitted in practice when using ket $|\ \rangle$ notation. 
This input tensor can be thought of as an MPS of bond dimension 1. 
As shown in Fig. \ref{fig:wav_mera}, this MPS becomes the bottom layer of a network
with wavelet MERA representing the upper layers. Each subsequent layer in the MERA is constructed by encoding wavelet transformations into the disentangler and isometry 
tensors $U$ and $V$. 

To accomplish such an encoding, we first decompose each of the wavelet coefficients 
in the set $\lbrace D_i \rbrace$ given in Eq.~\ref{eq:d4D} into two sequentially applied transformations. If we consider the term $x_i\ket{1}$ in each local feature vector 
$\ket{\phi(x_i)}$ of Eq.~\ref{psi_x} as a  ``particle'' whose state has a coefficient 
given by the input component $x_i$, we can trace the path of this particle 
through the MERA as shown in Fig.~\ref{fig:wav_mera}, 
assigning appropriate transformations to $x_i$ as it propagates through the tensors. 
Following this construction, one can work out the result of applying the MERA layer 
to a patch of four adjacent input tensors, whose dependence on input components
$(x_1,x_2,x_3,x_4)$ is to leading order:
\begin{align}
 \ket{\phi(x_1)} & \ket{\phi(x_2)}  \ket{\phi(x_3)}  \ket{\phi(x_4)} = \nonumber \\
= &\  (\ket{0} + x_1 \ket{1})\ (\ket{0} + x_2 \ket{1})\ \cdots  \nonumber \\ 
= &\ \ket{0}\ket{0}\ket{0}\ket{0} \nonumber \\
& + x_1 \ket{1}\ket{0}\ket{0}\ket{0} \nonumber \\
& + x_2 \ket{0}\ket{1}\ket{0}\ket{0} \nonumber \\
& + x_3 \ket{0}\ket{0}\ket{1}\ket{0}  \nonumber \\
& + x_4 \ket{0}\ket{0}\ket{0}\ket{1} + \ldots
\end{align}
where the omitted terms are higher-order in components of $\mathbf{x}$, such as $x_2 x_4 \ket{0}\ket{1}\ket{0}\ket{1}$. 

From the conditions shown in Fig.~\ref{fig:wav_mera}, it follows that the 
result of acting with the MERA layer on this patch of inputs is, to linear order
in the components of $\mathbf{x}$, an output vector
\begin{align}
(D_1 x_1 + D_2 x_2 + D_3 x_3 + D_4 x_4) \ket{1}
\end{align}
where now $\ket{1}$ labels the second basis vector of the vector space defined by the
tensor indices along the top of the MERA layer, and the coefficients $D_i$ are
related to the parameters in the MERA factor tensors as:

\begin{align}
D_1 & = -\sin{\theta_U}\cos{\theta_V} \\
D_2 & = \cos{\theta_U}\cos{\theta_V} \\
D_3 & = \cos{\theta_U}\sin{\theta_V} \\
D_4 & = \sin{\theta_U}\sin{\theta_V} \ .
\end{align} 
With the $D_i$ chosen to be the Daub4 wavelet coefficients Eq.~(\ref{eq:d4D}), 
this system of equations is easily solved by setting $\theta_U=\pi/6$ and $\theta_V=\pi/12$.
With this choice, the wavelets have successfully been encoded into the unitaries and isometries of our MERA.

While the above construction guarantees that the
action of a MERA layer on the input feature vector $\Phi(\mathbf{x})$ reproduces
the wavelet transformation for terms involving up to a single $\ket{1}$ basis vector,
such an exact correspondence no longer holds for terms with two or more $\ket{1}$
basis vectors in the tensor product. 
It is easy to show that when the $\ket{1}$ vectors are far enough apart,
the wavelet correspondence holds; when the $\ket{1}$ vectors are closer together, there
are corrections. However, we simply accept these as a defining property of our 
coarse-graining process, since choosing one wavelet family over another is already
somewhat arbitrary. It would be interesting in future work to explore how to make 
the mapping between wavelets and MERA layers more precise 
in the context of coarse-graining data.

Finally, we turn to how we coarse-grain each training sample through the MERA layers efficiently.
Recall that each sample is first converted to a rank-1 tensor (or product state in physics
terminology) of the form Eq.~(\ref{psi_x}), which we choose to view as an MPS tensor network of
bond dimension 1. Applying the first MERA layer to this MPS in a naive way would destroy the MPS form,
making any steps afterward very inefficient. However, we can proceed using a very accurate, controlled
approximation which is to apply the tensors in each MERA layer one by one, factoring the resulting
local tensors using a truncated SVD. This process is closely analogous to time evolution methods for
MPS such as time-evolving block decimation (TEBD) which are well developed in 
physics \cite{Paeckel:2019,Vidal:2004}.

\subsection{Training}

\begin{figure}[b]
\centering
\includegraphics[width=0.9\columnwidth]{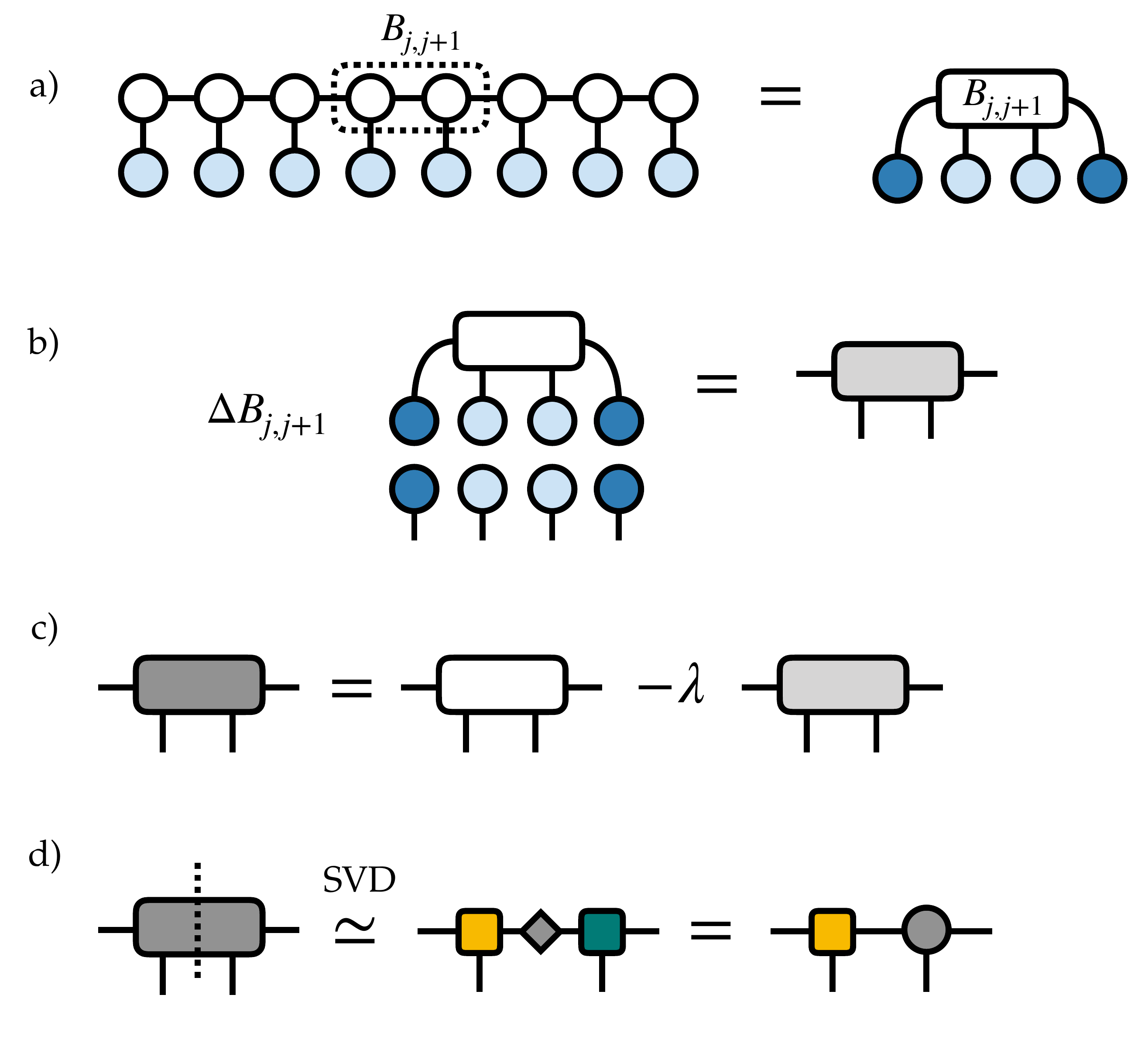}
\caption{The local update of $W^{(MPS)}$ is as follows: a) select a pair of neighboring $W^{(MPS)}$ tensors as $B_{jj+1}$, b) compute the gradient of the cost over all training examples, c) update $B_{jj+1}$, and d) decompose back into two MPS tensors using a truncated SVD. This is done for each pair in the MPS, sweeping back and forth until the cost is minimized.} 
\label{fig:mps_update}
\end{figure}

After each training data sample $\ket{\Phi(\mathbf{x}_i)}$ has been coarse grained through the MERA from size $N$ to $N' = N/2^L$ where $L$ is the number of wavelet-MERA layers used, 
we compute the (scalar) output of the model by an inner product with the tensor of weights $W$ at the 
topmost scale. We choose this weight tensor to be represented by an MPS:
\begin{equation}\label{Wmps}
W^{s_1 s_2 s_3 \ldots s_N} 
= \sum_{\{a\}} A^{s_1}_{a_1} A^{s_2}_{a_1 a_2} A^{s_3}_{a_2 a_3} \cdots A^{s_N}_{a_{N-1}}
\end{equation}
We then optimize the cost function in Eq. \ref{eq:cost} by sweeping back and forth through 
the tensors of the above MPS, updating each tensor in a DMRG-like fashion~\cite{StoudenmireJan2017}. 

The prominent feature of the optimization is a local update to $W$ by optimizing the MPS
tensors at sites $j$ and $j+1$ together. This is accomplished by constructing  
\begin{equation}
B^{s_j s_{j+1}}_{\alpha_{j-1}l\alpha_{j+1}} = A_{\alpha_{j-1}\alpha_j}^{s_j} A_{\alpha_j\alpha_{j+1}}^{s_{j+1}}   
\end{equation}   
Figure~\ref{fig:mps_update}(a) shows how the output of the model can be computed by first contracting
all of the coarse-grained data tensors with the MPS tensors not currently being optimized, then with
the bond tensor $B$. The gradient of the cost function is proportional to the tensor $\Delta B$ which
is shown in Fig.~\ref{fig:mps_update}(b). The $\Delta B$ tensor can be used to improve $B$ as shown in 
Fig.~\ref{fig:mps_update}(c). In practice, we actually use the conjugate gradient algorithm to perform
the optimization, but the first step of this algorithm is identical to Fig.~\ref{fig:mps_update}(b--c)
for the proper choice of $\lambda$. Finally, to proceed to the next step the MPS form of $W$ must
be restored to ensure efficiency. This can be done by treating $B$ as a matrix and computing its SVD
as shown in Fig.~\ref{fig:mps_update}(d). Truncating the smallest singular values of the SVD 
gives an adaptive way to automatically adjust the bond dimension of the weight MPS during training.

\subsection{Fine Scale Projection}

As a unique feature of our model, we introduce an additional step which provides the possibility of further optimizing the weight parameters for our classifier. 
Once training over the weight MPS has been completed at the topmost scale, the optimized weights can
be projected back through the MERA consisting of $N_{d4}$ layers to the previous scale defined by 
$N_{d4}-1$ layers.

\begin{figure}[t]
\centering
\includegraphics[width=\columnwidth]{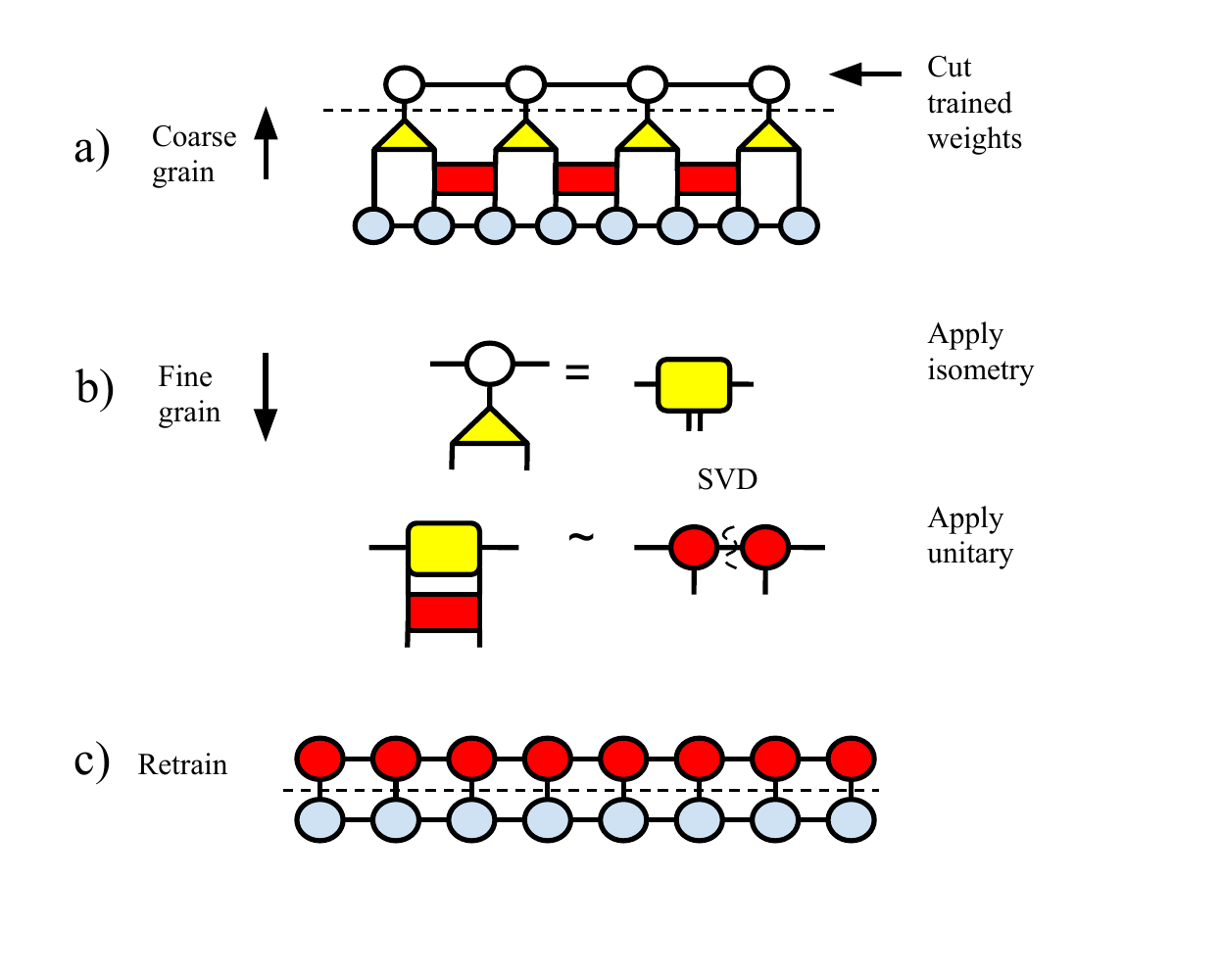}
\caption{a) The training sets are embedded into an MPS (light blue) and coarse grained through the wavelet MERA for training the weight MPS (white) at the top most layer. b) Once training is complete, the weight MPS is fine grained through the wavelet MERA by applying isometries and unitaries, and c) training is carried out on the training set at the finer scale} 
\label{fig:fine_projection}
\end{figure}

This is done by first applying the conjugate of the isometries of the topmost MERA layer 
to each MPS tensor representing $W$, 
thereby doubling the number of sites.
Next, we apply the conjugate of the unitary disentangler transformations to return to the 
basis defining the previous scale. 
Finally, to restore the MPS form of the transformed weights, we use an SVD factorization to split each
tensor so that the factors each carry one site or `feature' index. All of these steps are shown in 
Fig.~\ref{fig:fine_projection}(b).

The projection of these trained weights onto to the new finer scale serves as an initialization $W'$ 
for layer $N_{d4}-1$. At this step, 
the coarse-grained data previously stored at this finer scale are retrieved (having been previously saved in memory), and a new round of training is performed for $W'$. 
The intent is for this projected MPS $W'$ to provide a better initialization for the optimization than could have otherwise been
obtained directly at finer scales. Table~\ref{tab:algo} enumerates each step in our algorithm from initial coarse-graining to training and fine scale projection.

\begin{table}[t]\label{tab:algo}
\begin{center}
\caption{Algorithm}
\begin{tabular}{l|l}
\hline
\textbf{step 1} & Map each training sample $\mathbf{x}_i$ into $\ket{\Phi(\mathbf{x}_i)}$, Eq.~\ref{psi_x}\\
\hline
\textbf{step 2} & Coarse grain each $\ket{\Phi(\mathbf{x}_i)}$ through the wavelet MERA\\
\hline
\textbf{step 3} & Train $W$ at the current layer\\
\hline
\textbf{step 4} & Project the trained $W$ to the finer scale, Fig.~\ref{fig:fine_projection}\\
\hline
\textbf{step 5} & Repeat steps 3 and 4 as desired \\
\hline
\end{tabular}
\end{center}
\end{table} 

\section{Results}
\label{sec:results}

\subsection{DCASE}


The DCASE audio classification set consists of 15 batches (one batch per label), each containing 234 ten second audio clips for training ~\cite{dcase2017}. Each audio clip is a vector of 441,000 samples, which we embedded into a vector of $2^{19}$ elements by padding with zeros.
We constrained the problem to focus on binary classification, specifically distinguishing between "bus" and "beach" environment audio clips.

Each data set for the selected labels was coarse grained through $N_{h2}$ Haar transformation before encoding the data into our Daub4 wavelet MERA.  

\begin{figure}[t]
\centering
\includegraphics[width=\columnwidth]{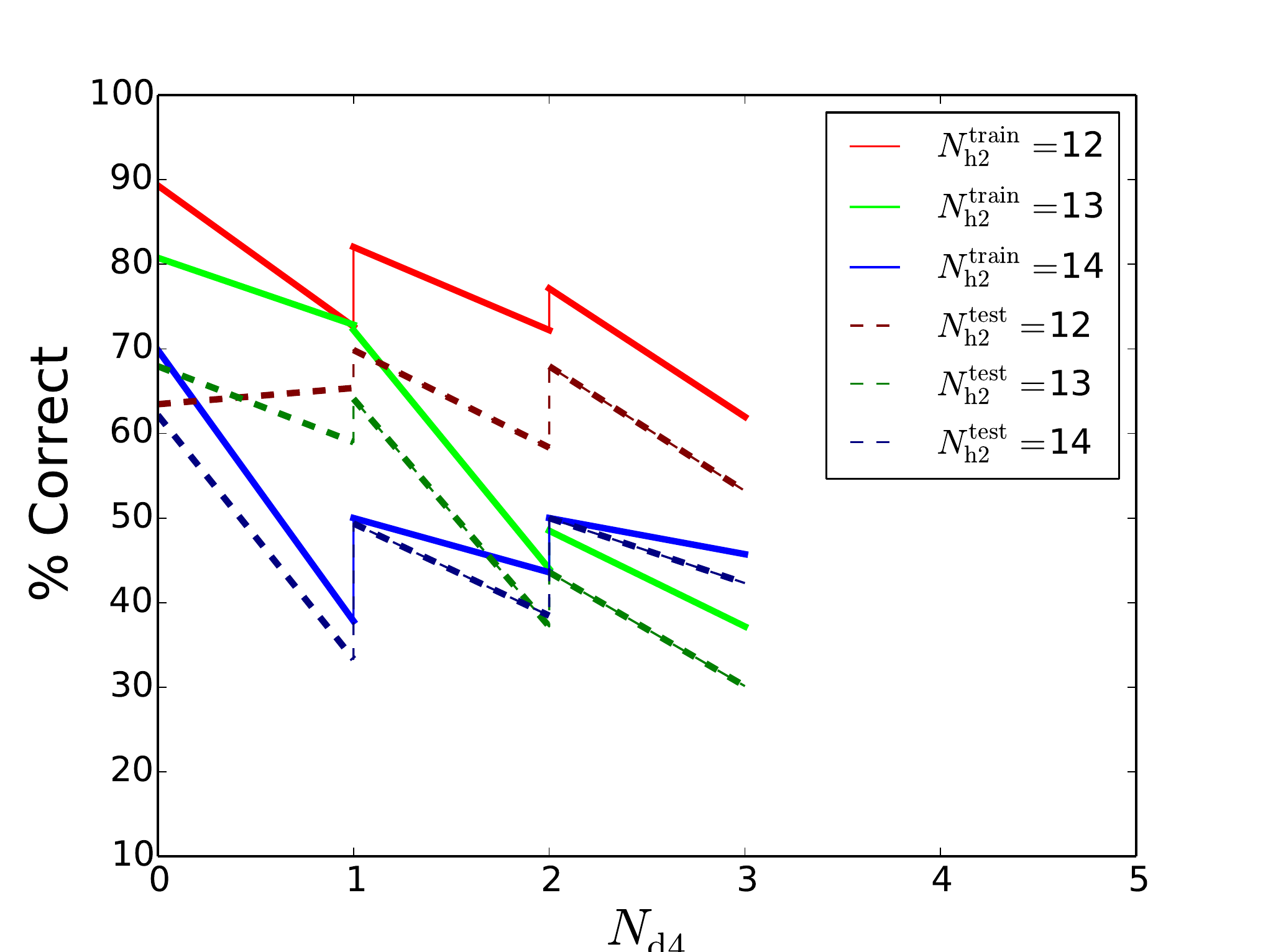}
\caption{The percentage of correctly labeled samples (training set as solid lines and test set as dashed lines) for the audio experiment after training $W$ over input data coarse grained through $N_{d4}+1$ Daub4 wavelet transformations and then trained again at $N_{d4}$ layers. Each data set was initially embedded into a vector of $2^{19}$ elements and coarse grained $N_{h2}$ Haar wavelet transformations before being embedded in the Daub4 wavelet MERA. The optimization was carried over 5 sweeps, keeping singular values above $\Delta = 1\times 10^{-14}$.} 
\label{fig:audio_data}
\end{figure}

We trained of the classifier at the top layer of the MERA for a 
varying number of  $N_{d4}+1$  Daub4 transformations, and subsequently projected the weights to one 
previous finer scale (scale $N_{d4}$). 
The percentage of correctly labeled examples after training at each scale
is shown in Fig.~\ref{fig:audio_data}. Each two point segment corresponds to training at the coarse scale (the right-most point in a segment), followed by training at the finer scale (the left-most point in a segment). 
The accuracy of the model decreases 
with the number of wavelet transformations. 
But it is also apparent that training the weights after $N_{d4}+1$ layers produces a better initialization and optimization for training at the $N_{d4}$ scale. Additionally, as can be noted by the closeness in accuracy between the training and testing sets for $N_{h2} = 14$, versus for $N_{h2}=12$, the generalization of the model improves with the number of wavelet layers applied. Optimization was carried over five sweeps, with the bond dimension of the weight MPS 
adaptively selected by keeping singular values above the threshold $\Delta = 1\times10^{-14}$.

\subsection{Regression}
\begin{figure}[b]
\centering
\includegraphics[width=\columnwidth]{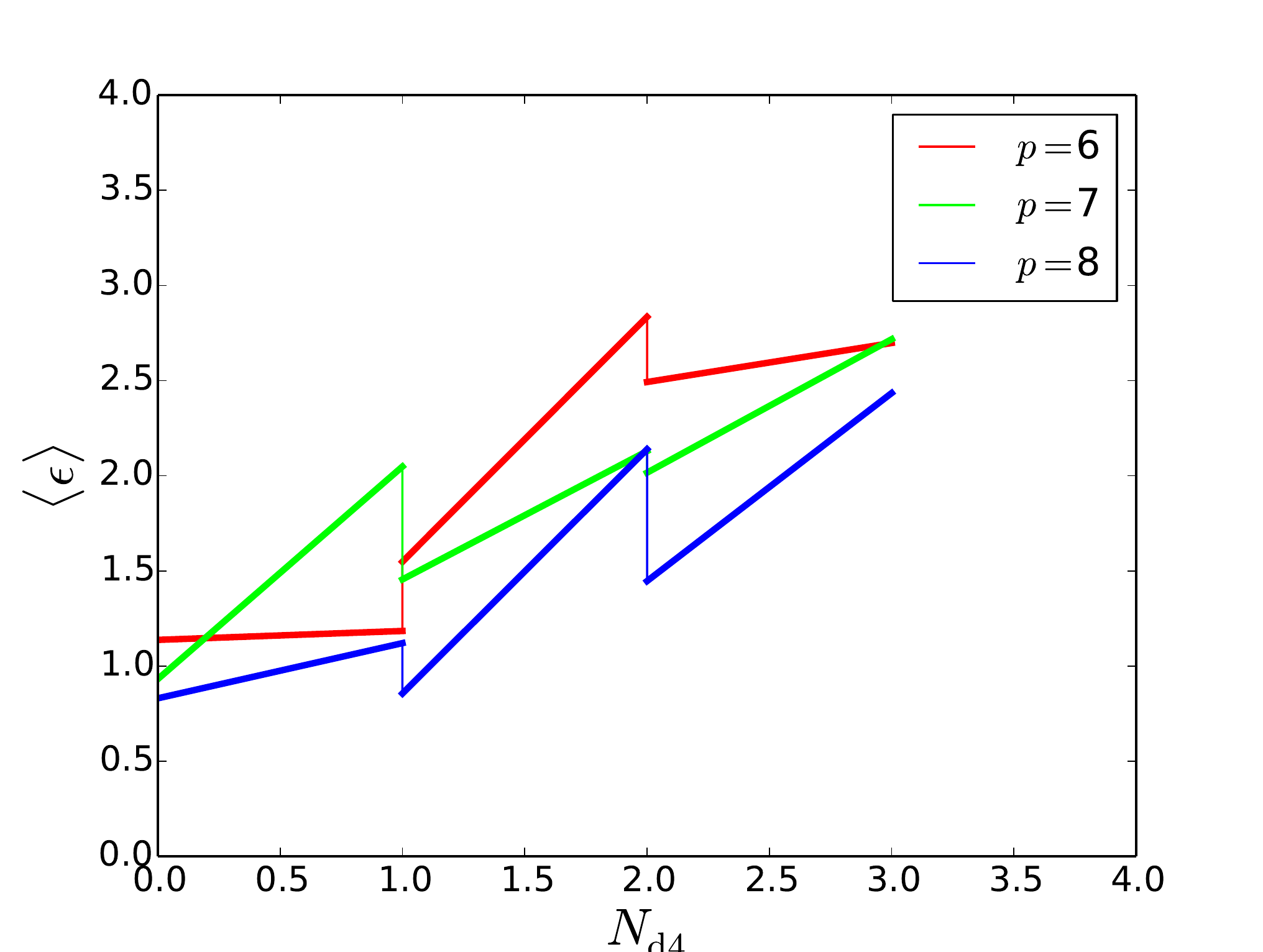}
    \caption{The average absolute deviation $\langle \epsilon \rangle$ (lower is better), for the temperature experiment after training $W^{(MPS)}$ over input data coarse grained through $N_{d4}+1$ wavelet layers and projected back to $N_{d4}$ layers. Each data set was initially constructed by selecting $p$ data points within the set $N_{fit}$. The optimization was carried over 40 sweeps, keeping singular values above $\Delta = 1\times 10^{-9}$.}  
\label{fig:weather_data}
\end{figure}

A single data file of the average daily temperatures of the Fisher River recorded from January 1, 1988 to December 31, 1991 were used to construct input data sets for regression in the following manner. By labeling each temperature $x_i$ for $0 < i < 1462$, the fitting interval $N_{fit}$ was taken as all the temperatures $\lbrace x_i | i \in [731,1461]\rbrace$. A single training example was constructed by selecting a contiguous block of $p$ temperatures from $N_{fit}$ as input for the MERA. The temperature immediately following this $p$ block was assigned as the label for that training example. By shifting the starting index of the $p$ block of temperatures, multiple examples could be constructed. In this way, the regression task was recast as a classification task over a continuous label. Because nearly every example contained a unique label, we used the average absolute difference of the interpolated label and the actual label, $\langle\epsilon\rangle$, as a measure of the accuracy of the model. The training phase of this data was carried out for forty sweeps, but with a singular value threshold set to $\Delta = 1\times 10^{-9}$. In Fig. \ref{fig:weather_data}, similar to the audio classification task, $\langle\epsilon\rangle$ is given for input data coarse-grained through $N_{d4}+1$, with the weights trained at this scale, and then projected onto the finer scale $N_{d4}$ for further training. Each two-segment section is representative of training at $N_{d4}+1$ layers on the right-most point and training at $N_{d4}$ layers on the left-most point. Again, we note a general decrease in accuracy with the number of wavelet transformations, but an improvement in accuracy when the weights are initialized by optimization through $N_{d4}+1$ wavelet layers, as compared to directly  randomly initializing at $N_{d4}$ layers. 

\section{Discussion \label{sec:discussion}}

MERA are a family of tensor network factorizations which 
process information in a hierarchical way and preserve computational advantages associated
with tree-tensor networks, such as efficient marginalization, yet can be more expressive and 
powerful because of extra ``disentangler'' layers which mix branches of the tree.
Here we have proposed an architecture for machine learning which uses MERA layers which
approximate wavelet scaling functions as a preprocessing
step, and explored the advantages of this choice. Because of the presence of disentanglers, MERA
are able to approximate non-trivial families of wavelets with overlapping support,
such as Daubechies-4 wavelets. 

The fact that a MERA is a tensor network composed of multi-linear 
transformations allows our model to have an interesting reversibility property: 
the trainable parameters of the model can be projected to a finer scale of resolution while 
preserving the output of the model, allowing this procedure to initialize more expressive models 
by initially training models with fewer parameters. This initialization step showed notable improvement in the accuracy of the model as compared to the direct initialization of the weights at any given scale, for a fixed
number of optimization sweeps at the finer scale.

Among techniques widely used in machine learning, our model architecture most closely resembles 
a convolutional neural network (CNN) \cite{Schmidhuber2015}. 
In both architectures, data is initially processed through
a set of layers which mix information in a locality-preserving way.
Pooling layers commonly used in CNNs closely resemble the isometry maps in a MERA, which act as
a coarse-graining transformation.
The connection to CNNs suggests one future direction we could explore would be to make the 
parameters of the MERA layers in our model adjustable, and train them along with the weights 
in the top layer. More ambitiously, instead of just training the parameters in a given MERA layer,
one can envision computing this layer from a set of weights at a given scale through a controlled
factorization procedure.

By noting the dependence of the accuracy of our model on the number of wavelet layers, we deduce that the input data can exhibit correlations at length scales that can be lost through the wavelet transformations. It is therefore important to tailor the number of wavelets and initial size to the specific data set being analyzed in order to maintain a desirable accuracy. Our setup gives an affordable and adaptive way to strike a balance
between the efficiency and generalization gains obtained by coarse-graining
versus model expressivity by trading off one for the other through the fine-graining procedure.
One way to select the best number of layers to use would be to use too many intentionally, then fine-grain
until the gains in model performance begin to saturate, or until generalization starts to degrade.
 
We conclude that our algorithm provides an interesting platform for classification and regression, with
unique capabilities. Further work is needed to improve the model accuracy for the classification of continuous labels (i.e. regression). It is also worth investigating the effect that different wavelet transformations may have on the model; determining how much is gained by introducing optimizable parameters into the coarse-graining layers; and investigating adaptive learning schemes for the sizes of indices in the MERA layers.

\acknowledgments

J.R. was partially supported by NSF Grants No. CCF-1525943 and CCF-1844434.
E.M.S is supported by the Flatiron Institute, a division of the Simons Foundation.
\bibliographystyle{unsrt}
\bibliography{biblio}

\end{document}